\documentclass[conference]{IEEEtran}
\IEEEoverridecommandlockouts
\usepackage{etoolbox}
\makeatletter
\patchcmd{\@makecaption}
  {\scshape}
  {}
  {}
  {}
\makeatother
\usepackage{threeparttable, booktabs}
\usepackage{cite}
\usepackage{amsmath,amssymb,amsfonts}
\usepackage{algorithmic}
\usepackage{algorithm}
\usepackage{graphicx}
\usepackage{booktabs}
\usepackage{textcomp}
\usepackage{hyperref}
\usepackage[section]{placeins}
\def\BibTeX{{\rm B\kern-.05em{\sc i\kern-.025em b}\kern-.08em
    T\kern-.1667em\lower.7ex\hbox{E}\kern-.125emX}}
\begin{document}

\title{Local intrinsic dimensionality estimators based on concentration of measure\\
\thanks{This work was funded in part by the French government under management of Agence Nationale de la Recherche as part of the "Investissements d'avenir" program, reference ANR-19-P3IA-0001 (PRAIRIE 3IA Institute), the Ministry of education and science of Russia (Project No. 14.Y26.31.0022), the Association Science et Technologie, the Institut de Recherches Internationales Servier and the doctoral school Frontières de l'Innovation en Recherche et Education–Programme Bettencourt.
}}

\author{\IEEEauthorblockN{Jonathan Bac}
\IEEEauthorblockA{
\textit{Institut Curie, PSL Research University,}\\
\textit{Mines ParisTech, INSERM, U900}\\
\textit{Centre de Recherches Interdisciplinaires,}\\
Paris, France\\
jonathan.bac@cri-paris.org}
\and
\IEEEauthorblockN{Andrei Zinovyev}
\IEEEauthorblockA{
\textit{Institut Curie, PSL Research University,}\\
\textit{Mines ParisTech, INSERM, U900}\\
Paris, France \\
\textit{Lobachevsky University}
\\
Nizhni Novgorod, Russia \\
andrei.zinovyev@curie.fr} \\

}

\maketitle

\begin{abstract}
Intrinsic dimensionality (ID) is one of the most fundamental characteristics of multi-dimensional data point clouds. Knowing ID is crucial to choose the appropriate machine learning approach as well as to understand its behavior and validate it. ID can be computed globally for the whole data point distribution, or computed locally in different regions of the data space. In this paper, we introduce new local estimators of ID based on linear separability of multi-dimensional data point clouds, which is one of the manifestations of concentration of measure. We empirically study the properties of these estimators and compare them with other recently introduced ID estimators exploiting various effects of measure concentration. Observed differences between estimators can be used to anticipate their behaviour in practical applications.
\end{abstract}

\begin{IEEEkeywords}
high-dimensional data, intrinsic dimension, effective dimension, concentration of measure, linear separability
\end{IEEEkeywords}

\section{Introduction}

Datasets used in applications of machine learning frequently contain objects characterized by thousands or millions of features. In this respect, the well-known \textit{curse of dimensionality} is frequently discussed which states that many problems become exponentially difficult in high dimensions \cite{Gorban2020}. However, in what concerns the application of machine learning methods, the curse of dimensionality is not automatically manifested when the number of features is large: this depends rather on the dataset's intrinsic dimensionality (ID). If the features of a dataset are correlated in linear or non-linear fashion then the data point cloud can be located close to a subspace of relatively low ID. This makes appropriate the application of dimension reduction algorithms to obtain a lower-dimensional representation of data. By contrast, if the value of ID is high then the data point cloud becomes sparse and its geometrical and topological properties can be highly non-intuitive, in particular, due to various manifestations of concentration of measure \cite{Gromov2003}. 

Therefore, the estimation of ID is crucial to the choice of machine learning methodology and its applications, including validation, explainability and deployment. Indeed, ID determines to a large extent the strategy and feasibility of validating an algorithm, estimating the uncertainty of its predictions and explaining its decisions. Such explanations can even be legally required for its use in sensitive applications \cite{jia2019improving}. 

There has been recent progress on these topics, e.g. with several approaches to explain and validate classifiers by looking at various data properties related to ID \cite{ribeiro2016should,jiang2018trust,jia2019improving}. ID can pose fundamental limits on the robustness of classifiers; this has been illustrated recently with adversarial examples, where a minimal perturbation to the input can lead to a misclassification. Recent theoretical results have shown that such adversarial examples are inevitable for a concentrated metric probability space \cite{gilmer2018adversarial,fawzi2018adversarial,bhagoji2019lower} and algorithms have been proposed that estimate adversarial risk by quantifying concentration \cite{mahloujifar2019empirically}. 

From a more general perspective, theory and algorithms for correcting artificial intelligence-based systems have been developed that exploit properties specific to spaces possessing large ID \cite{GORBAN2018303,Tyukin2019}. In this respect, a complementarity principle has been formulated \cite{Gorban2018Blessing,Symphony2019,Bac2020}: the data space can be split into a low volume (low dimensional) subset, where nonlinear methods are effective, and a high-dimensional subset, characterized by measure concentration and simplicity, allowing the effective application of linear methods.

One important observation is that in real life the ID of the complete dataset might not be equal to the ID of its parts. Therefore, ID can be considered a local characteristic of the data space, defined in each data neighborhood. In this case we refer to it as \textit{local intrinsic dimensionality}. Even when there are no ID variations in a dataset, the relation between global and local intrinsic dimensionality can be non-trivial - for example, one can easily construct examples of datasets which are low-dimensional locally but globally possess large linear dimension \cite{Bac2020}.

Methods for estimating global ID can be applied locally in a data neighbourhood, and, vice versa, local ID estimators can be used to derive a global ID estimate. However, local ID estimation presents specific challenges, such as dealing with restricted cardinality, which render many global methods ineffective in practice. These challenges explain the rise in recent works dedicated specifically to developing local ID estimators \cite{amsaleg2019intrinsic,chelly2016enhanced,Kerstin,SSV,ANOVA,albergante2019estimating,erba2019intrinsic}.

Well-known methods for estimating global or local ID are based on Principal Component Analysis (PCA) and quantifications of the covariance matrix's eigenspectrum using various heuristics (thresholding total explained variance, limiting the conditional number, using reference spectra such as broken stick distribution, etc.) \cite{Fukunaga1971,bruske1998intrinsic,Fan,little2009estimation}. Other famous examples include maximum likelihood estimation (MLE) and the correlation dimension, based on counting the number of objects in a growing neighbourhood \cite{Grassberger1983,Levina2004}. Since PCA is a linear method, it generally tends to overestimate ID, while MLE and correlation dimension both tend to underestimate larger ID values.

In order to overcome these limitations, several new ID estimators have been recently introduced that exploit concentration of measure \cite{Gorban2018Blessing,albergante2019estimating,DANCo,Kerstin,ANOVA,Facco2017a,SSV}. Increasing ID results in various manifestations of measure concentration that scale differently with respect to dimensionality. For example, linear separability of points increases rapidly, such that 30 dimensions can be already considered large, while the appearance of exponentially large quasi-orthogonal bases or hubness are usually manifested in higher dimensions \cite{Gorban2016,Gorban2020}. These differences lead to different properties of ID estimators, e.g. sensitivity to different dimensions and dependence of the estimated ID on sample size. 


As noted in \cite{SSV}, most of the local ID estimators use pairwise distances (relationship between two points) or angular information (relationship between three points) \cite{Levina2004,DANCo,ANOVA}, whereas others take into account the volume of a $d$-simplex (i.e., the relationship between $d+1$ points) \cite{SSV,Kerstin} or in our case, the separability probability of each point from the others \cite{albergante2019estimating}. The common approach is to assume local distributions of data points are close to a uniformly sampled unit $n$-ball $B^{n}$ or their scaled vectors to a uniformly sampled unit sphere $S^{n-1}$. Then, various sample statistics are used whose dependence on $n$ is theoretically established for uniform distributions on $B^{n}$ or $S^{n-1}$. If the estimated statistics of a data sample are similar to the theoretical of $B^{n}$ or $S^{n-1}$ then the dimensionality of the sample is estimated to be $n$. 

For example, DANCo \cite{DANCo} estimates the probability density function of normalized nearest neighbor distance from the center (exploiting concentration of norms), and the parameters of a Von-Mises distribution (exploiting concentration of angles). Expected Simplex Skewness (ESS) \cite{Kerstin}, in its default version, computes simplex skewness, defined as the ratio between the volume of a simplex with one vertex in the centroid and the others in data points, and the volume this simplex would have if edges incident to the centroid were orthogonal. For a $d$-simplex with $d=1$, skewness is $sin(\theta)$, with $\theta$ the angle between the two edges incident to the centroid vertex. The mean sample skewness is compared to the ESS for uniformly distributed data on $B^{n}$.
ANOVA \cite{ANOVA} uses a $U$-statistic for the variance of the angle between pairs of vectors among uniformly chosen points in $S^{n-1}$. 

In our previous work we introduced and benchmarked global estimators of ID based on Fisher separability, using theoretical results by Gorban, Tyukin et al. \cite{GORBAN2018303,albergante2019estimating}. The details of this approach are provided below. It appeared that for noisy samples from synthetic manifolds, the method was competitive with other ID estimators. In particular, its behavior was close to two recently introduced ID estimators based on concentration of measure, namely ESS and DANCo. 

In this work we extend our previous study and introduce two ways to estimate ID locally, in each data point, based on Fisher separability properties. We compare the properties of these local dimensionality estimators with other estimators based on quantifying various manifestations of concentration of measure. 

\section{Local ID estimation based on Fisher separability}

In the present work, we will follow the notations introduced in the works by A.Gorban, I.Tyukin and their colleagues \cite{GORBAN2018303}: we call a data vector $\mathbf{x}\in R^n$ \textit{linearly separable} from a finite set of points $Y \subset R^n$ if there exists a linear functional $l$ such that $l(\mathbf{x})>l(\mathbf{y})$ for all $\mathbf{y} \in Y$. If for any point $\mathbf{x}$ there exists a linear functional separating it from all other data points, then such a data point cloud is called {\it linearly separable} or $1$-convex. The separating functional $l$ may be computed using the linear Support Vector Machine (SVM) algorithms, the Rosenblatt perceptron algorithm, or other comparable methods. However, these computations may be rather costly for large-scale estimates. Hence, in the works of Gorban, Tyukin and their colleagues it was suggested to use a non-iterative estimate of the linear functional using Fisher's discriminant which is computationally inexpensive, after a standard pre-processing: \cite{GORBAN2018303}.

\begin{enumerate}
    \item centering
    \item performing linear dimensionality reduction by projecting the dataset into the space of $k$ first principal components, where $k$ may be relatively large. In practice, we select the largest $k$ (in their natural ranking) such that the corresponding eigenvalue $\lambda_k$ is not smaller that $\lambda_1/C$, where $C$ is a predefined threshold. Under most circumstances, $C = 10$
    \item whitening (i.e., applying a linear transformation after which the covariance matrix becomes the identity matrix)
\end{enumerate}

After such normalization of $X$, it is said that a point $\mathbf{x} \in X$ is Fisher-linearly separable from the cloud of points $Y$ with parameter $\alpha$, if

\begin{equation*}\label{onepoint_separability_criterion}
(\mathbf{x},\mathbf{y}) \leq \alpha (\mathbf{x},\mathbf{x})
\end{equation*}

\noindent for all $\mathbf{y} \in Y$, where $\alpha \in \left[0,1\right)$. If equation (\ref{onepoint_separability_criterion}) is valid for each point $\mathbf{x} \in X$ such that $Y$ is the set of points $\mathbf{y} \neq \mathbf{x}$ then we call the dataset $X$ Fisher-separable with parameter $\alpha$. In order to quantify deviation from perfect separability, let us introduce $p_\alpha(\mathbf{y})$, the probability that the point $\mathbf{y}$  is inseparable from all other points. 

In order to associate a value of ID to a point $\mathbf{y}$ in the data space, we compare the empirical $p_\alpha(\mathbf{y})$ estimates to the $p_\alpha$ of some reference data distribution whose dimension is known and separability properties can be analytically derived. The simplest such distribution is the uniform distribution of vectors on the surface of a unit $n$-dimensional sphere. Since this distribution is uniform, $p_\alpha$ does not depend on a data point and equals in any point (see derivation in \cite{GORBAN2018303,Gorban2020}): 

\begin{equation}\label{UnitSphereDistribution}
p_\alpha \lessapprox \frac{(1-\alpha^2)^{\frac{n-1}{2}}}{\alpha\sqrt{2\pi n}}
\end{equation}

By resolving this formula with respect to $n$, we derive the following value of dimensionality as a function of inseparability probability, for a uniform distribution on the surface of the unit $n$-dimensional sphere:

\begin{equation}\label{ComputingEffectiveDimension}
n_\alpha = \frac{W(\frac{-\ln(1-\alpha^2)}{2\pi p_\alpha^2\alpha^2(1-\alpha^2)})}{-\ln(1-\alpha^2)}
\end{equation}

\noindent where $W(x)$ is the real-valued branch of the Lambert W function. As a reminder, the Lambert W function solves equation $v = we^w$ with respect to $w$, i.e. $w=W(v)$. The pseudo-code of the algorithm for computing $n_\alpha$ was provided by us earlier \cite{albergante2019estimating}.

Here we should make several important notes. Firstly, in order to apply the ID estimate (\ref{ComputingEffectiveDimension}) to a dataset $X$, one should apply an additional scaling step besides the preprocessing steps (1)-(3) described above. The scaling consists in normalizing each vector to the unit length, which corresponds to the projection onto a unit sphere. It means that in practice we do not distinguish an $n$-ball $B^{n}$ from $n$-sphere $S^{n-1}$, both give ID=$n$ in our case. This can lead to shifting by value 1 the estimations of ID in small dimensions, especially in artificial benchmark examples.

Secondly, as mentioned in \cite{Gorban2020}, the formula (\ref{UnitSphereDistribution}) is an estimation from above, meaning that the actual empirical value of $p_\alpha$ is strictly less than the right hand side of (\ref{UnitSphereDistribution}). In particular, for some points, data point density, and values of $\alpha$ one can have empirical estimate $p_\alpha=0$ which makes (\ref{ComputingEffectiveDimension}) inapplicable. The value of $\alpha$ should be adjusted in order to avoid the mean of $p_\alpha$ being too close to zero and in order to avoid too strong finite sampling effects (see \cite{albergante2019estimating} and Figure~\ref{fig1}). This consideration also provides some theoretical limits on the maximally detectable dimensionalities, as shown hereafter in Figure 2.

Based on the above definitions, the separability properties of  the data point cloud can be globally characterized by the histogram of empirical $p_\alpha$ distribution (probabilities of individual point inseparability) and the profile of intrinsic dimensions $n_\alpha$ (\ref{ComputingEffectiveDimension})
for a range of $\alpha$ values (e.g., $\alpha \in [0.6,...,1.0]$).

Let us denote $\bar{p}_\alpha(\mathbf{X})$ the mean value of the distribution of $p_\alpha(\mathbf{x})$ over all data points. We can introduce the global estimate of ID as in our previous work (\cite{albergante2019estimating}):

\begin{equation}\label{GlobalID}
n^{global}_\alpha = \frac{W(\frac{-\ln(1-\alpha^2)}{2\pi \bar{p}_\alpha^2\alpha^2(1-\alpha^2)})}{-\ln(1-\alpha^2)}.
\end{equation}

Now let us specify two local ID estimates based on Fisher separability. The first one will simply use the formula (\ref{ComputingEffectiveDimension}) in order to estimate $n_\alpha$ in a data point $\mathbf{y}$. We will call this estimate \textit{global pointwise intrinsic dimension}, since in its definition the global separability properties of $X$ are exploited, but the ID is computed in a data point:

\begin{equation}\label{GlobalPointwiseID}
n_\alpha(y) = \frac{W(\frac{-\ln(1-\alpha^2)}{2\pi p_\alpha(y)^2\alpha^2(1-\alpha^2)})}{-\ln(1-\alpha^2)}.
\end{equation}

At the same time we can follow the standard approach of defining data neighborhoods and compute the $n^{global}_\alpha$ for these fragments of data. One of the simplest way to define a local neighbourhood is to determine the first $k$ nearest neighbours (kNN) for a data point. This naive approach has well-known drawbacks, such as the edge effect : the ratio between points close to the border of the manifold and points inside it increases with dimensionality, meaning kNN data neighbourhoods can deviate from uniform distributions. \cite{DANCo,verveer1995evaluation}. Forming kNN using distances of the original space can also create neighborhoods that do not reflect geodesic distance on the manifold. Recent work has suggested new ways to tackle these issues \cite{amsaleg2019intrinsic, chelly2016enhanced}. Nonetheless this basic approach provides an easy way to start applying methods locally. Therefore, we define \textit{local kNN ID} as:

\begin{equation}\label{LocalkNNID}
n^{kNN}_\alpha(y) = \frac{W(\frac{-\ln(1-\alpha^2)}{2\pi p^{kNN}_\alpha(y)^2\alpha^2(1-\alpha^2)})}{-\ln(1-\alpha^2)},
\end{equation}

\noindent where $p^{kNN}_\alpha(y)$ is computed for a dataset comprising $k$ nearest neighbours of $y$. In practical applications, $k$ is chosen of few hundreds by the order of magnitude.

\section{Numerical results}

\subsection{Estimation of ID for $n$-dimensional balls} 

\begin{figure*}[!htb]
{\centering 
\includegraphics[width=\textwidth,height=\textheight,keepaspectratio]{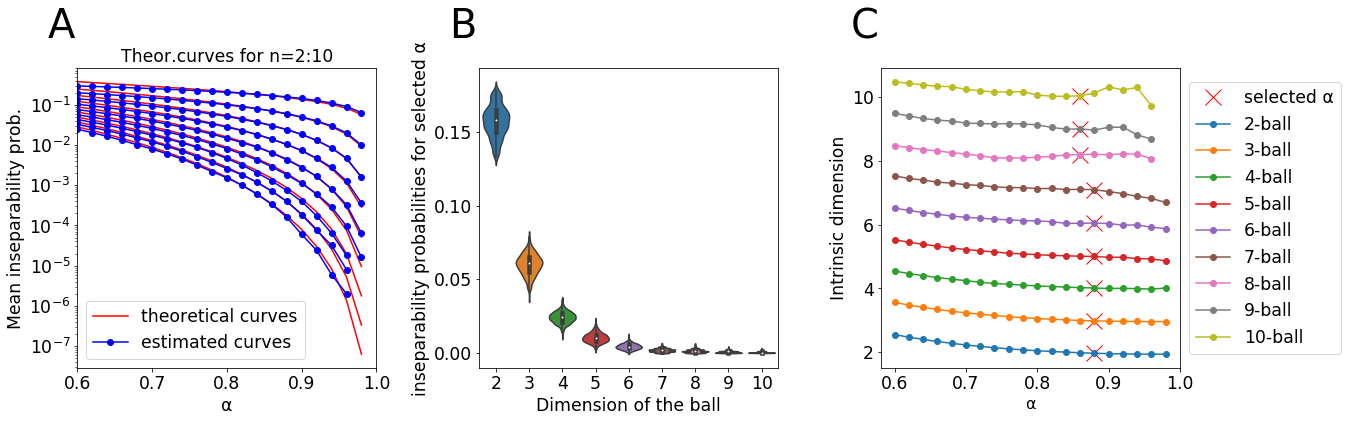}
\includegraphics[width=\textwidth,height=\textheight,keepaspectratio]{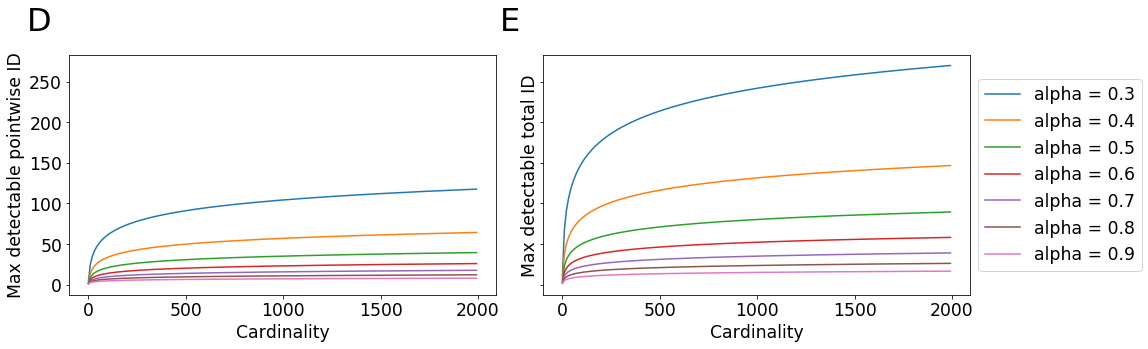}
\includegraphics[width=\textwidth,height=\textheight,keepaspectratio]{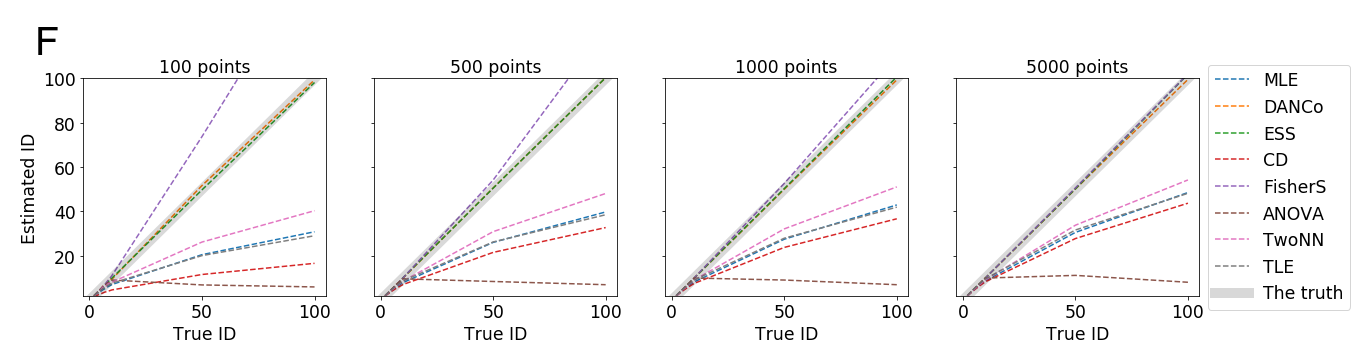}
\includegraphics[width=\textwidth,height=\textheight,keepaspectratio]{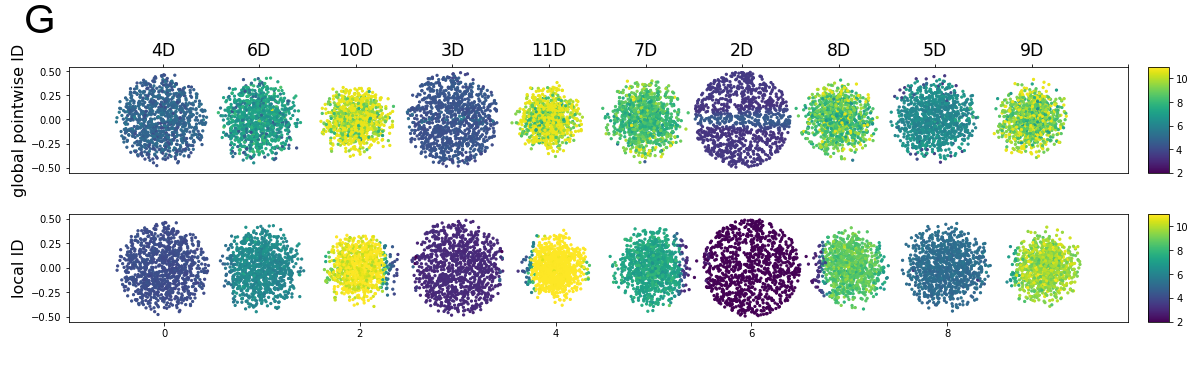}
\caption{\textbf{A-} theoretical vs estimated mean inseparability probability for balls of increasing dimension; \textbf{B-} complete inseparability probability histograms for selected $\alpha=.88$; \textbf{C-} intrinsic dimension estimated from the mean inseparability probability. Due to measure concentration, inseparability probability sharply decreases as a function of dimension. \textbf{D-} maximum detectable pointwise intrinsic dimension; \textbf{E-} maximum detectable "total" ID (estimate from the mean inseparability of all data points); \textbf{F-} ability of various estimators to quantify high intrinsic dimensionality from different sample size, tested on the simplest $n$-ball example. Here CD stands for the Correlation Dimension, FisherS is the global ID estimation based on Fisher separability, MLE stands for Maximum Likelihood Estimation of Intrinsic Dimension \cite{Levina2004} and TLE stands for Tight Locality Estimator \cite{amsaleg2019intrinsic}. Other estimator names are referenced in the Introduction;
\textbf{G-} local and global pointwise ID estimates computed on a synthetic dataset containing 10 unit balls of different dimensions (2D to 11D, each sampled with 500 points) arranged in a line. Maximum ID is capped to the embedding dimension (11)}.

\label{fig1}}
\end{figure*}

We illustrate our approach for a simple example of $n$-ball $B^n$ (see Figure~\ref{fig1}), sampled by 2000 points. The empirically estimated values of $p_\alpha$ are shown together with the theoretical dependence (\ref{UnitSphereDistribution}) in Figure~\ref{fig1}A for $n=2...10$. Figure~\ref{fig1}B shows the empirically estimated distributions of $p_\alpha$ for $n=2...10$. In Figure~\ref{fig1}C we show the empirically estimated value of $n$ as a function of $\alpha$. One can see that smaller values of $\alpha$ create biased estimates of $n$. At the same time, maximum possible values of $\alpha$ (for which the mean of $p_\alpha$ is still non-zero) create unstable estimates suffering from finite sampling effects, starting from $n=7$. Therefore, for the unique definition of $n$, our rule of thumb is to select $p_\alpha$ which would equal 0.8 multiplied by the maximally measured $p_\alpha$ (shown by crosses in Figure~\ref{fig1}C). 

Finite sample effects pose theoretical limits on maximum measurable values of ID from separability analysis as a function of the number of points $N$. The minimum measurable $p_\alpha$ equals $\frac{1}{N}$ (here we neglect small difference between $N$ and $N-1$): therefore, maximally measurable $n_\alpha(y)$ (\ref{GlobalPointwiseID}) equals $\frac{W(\frac{-\ln(1-\alpha^2)N^2}{2\pi \alpha^2(1-\alpha^2)})}{-\ln(1-\alpha^2)}$ (shown in Figure~\ref{fig1}D). At the same time, the mean of $p_\alpha$ is non-zero except if the dataset is completely Fisher-separable. This defines the limit for measuring (\ref{GlobalID}) in $\frac{W(\frac{-\ln(1-\alpha^2)N^4}{2\pi \alpha^2(1-\alpha^2)})}{-\ln(1-\alpha^2)}$ (shown in Figure~\ref{fig1}E) since the minimally measurable 
$\bar{p}_\alpha$ can be estimated as $\frac{1}{N^2}$. This conclusion allows us to estimate the maximally measurable $n^{kNN}_\alpha(y)$ (\ref{LocalkNNID}) in $\frac{W(\frac{-\ln(1-\alpha^2)k^4}{2\pi \alpha^2(1-\alpha^2)})}{-\ln(1-\alpha^2)}$ which can be read from Figure~\ref{fig1}E for cardinality equals $k$. Notice from Figure~\ref{fig1}D,E that the maximum measurable ID quickly saturates with the number of points $N$ but remains relatively high for smaller $\alpha$. 


For the $n$-ball example, we also studied how well different estimators can characterize the ID of high dimensional samples, depending on their cardinality (see Figure~\ref{fig1}F). In the case of this simple benchmark, ESS and DANCo show impressive results, giving the exact value of ID even for very small samples (ESS is even able to estimate dimensionality higher than the number of points, which seems to be counter-intuitive at first glance but this is discussed in the paper \cite{Kerstin}). Fisher separability analysis tends to overestimate the ID in case of small sample sizes but works as well as DANCo and ESS for larger number of points. The overestimation is connected to the point previously discussed of maximally measurable dimensionality for a given $\alpha$ value. In case of a small sample in very high dimension, the value of $\alpha$, adjusted using the heuristics described above, becomes small in order to avoid full separability: in this case, the estimate (\ref{GlobalID}) becomes less accurate. 

We also note that many widely used ID estimators such as the correlation dimension or MLE heavily ID, showing saturation at few tens of dimensions. The ANOVA estimator showed surprising underestimation of larger ID even for large number of points (Figure~\ref{fig1}F).

\subsection{Synthetic and real-life examples}

In Figure~\ref{fig1}G we show visualizations of ID values for a simple synthetic '10 balls' dataset, which represents 10 unit $n$-balls of increasing $n=2..11$ embedded without intersection in 11-dimensional space such that the $n$-ball shares  $n-1$ dimensions with the $(n-1)$-ball. The balls are separated by increments of 1 along the first axis. From this example, one can see that both pointwise global and local kNN ID based on Fisher separability characterize well the dimensionality of the balls, however, pointwise global estimates are more heterogeneous within a ball while the local kNN estimate suffers when the data neighborhood includes points from different balls. 

\begin{figure*}[!htb]
{\centering 
\includegraphics[width=.95\textwidth,height=.95\textheight,keepaspectratio,trim={0 2.8cm 0 0},clip]{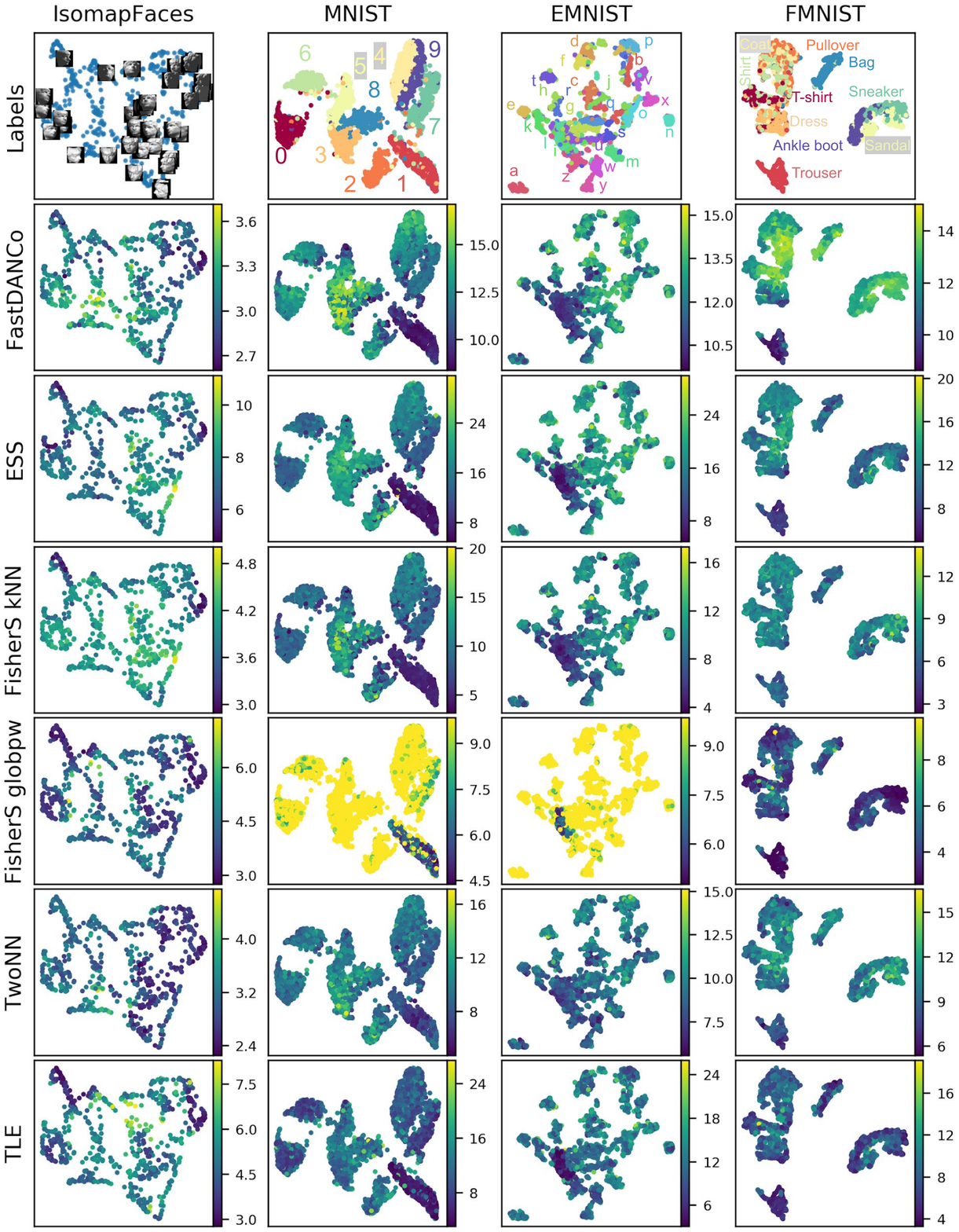}
\caption{Visualization of local ID measures (obtained with the 100 nearest neighbors of each point) on 2D layouts of real datasets projected onto their first 100 principal components and subsampled to 5000 points. For the  MNIST datasets, the class labels are annotated directly on the UMAP scatter plot with colored labels located close to the corresponding clusters.}
\label{fig2}}
\end{figure*}

In Figure~\ref{fig2}, we visualize local ID values from several ID estimators on real-life datasets. A first observation is that local ID can be highly variable from one data cluster to another. For example, in the MNIST datasets, local ID highlights some clusters which possess much lower local ID than the others (digit '1' - for the MNIST digit dataset, letters 'i' and 'l' for the MNIST letters dataset, and cluster 'Trousers' for the fashion MNIST dataset). This might reflect the intrinsic number of degrees of freedom in the distribution of different realizations (e.g., writing or design) of the same object type (e.g., a digit). At the same time, in the case of the ISOMAP Faces dataset, the distribution of local ID is more uniform and close to 3 as expected, although there is variation. This dataset is more challenging for local ID estimation since it only contains ~700 data points. ESS and TLE in particular seem to overestimate ID. 

The synthetic balls dataset (Figure~\ref{fig1}G) and real data examples (Figure~\ref{fig2}) demonstrate possible relations between global pointwise and local (kNN) ID estimates based on Fisher separability. Global pointwise ID estimates are computed based on the separability of a point from the rest of the data point cloud, while local kNN ID estimates are based on the mean inseparability of points in each local neighbourhood. These two estimates do not have to be similar: strong deviations of one from another can indicate complex data topology, when distant parts of the data point cloud are co-localized in close data subspaces. Local inseparability usually imposes global inseparability but not vice versa. Global pointwise ID can thus provide insights about the structure of the point cloud but will usually give less accurate ID estimates.


\subsection{Dependence of ID estimators on the subsample size}

A desirable property of an ID estimator is the ability to quantify a wide range of ID values from a relatively small sample size. In order to study these properties, we took several synthetic and real-life datasets and computed global ID and local ID with one hundred nearest neighbors for five estimators, on different subsample sizes. Running these estimations 10 times also gave us an idea about the uncertainty in the estimates for a given subsample size. 

\begin{figure*}[!htb]
{\centering 
\includegraphics[trim={1cm 5cm 1cm 2.1cm}, clip,width=\textwidth,height=\textheight,keepaspectratio]{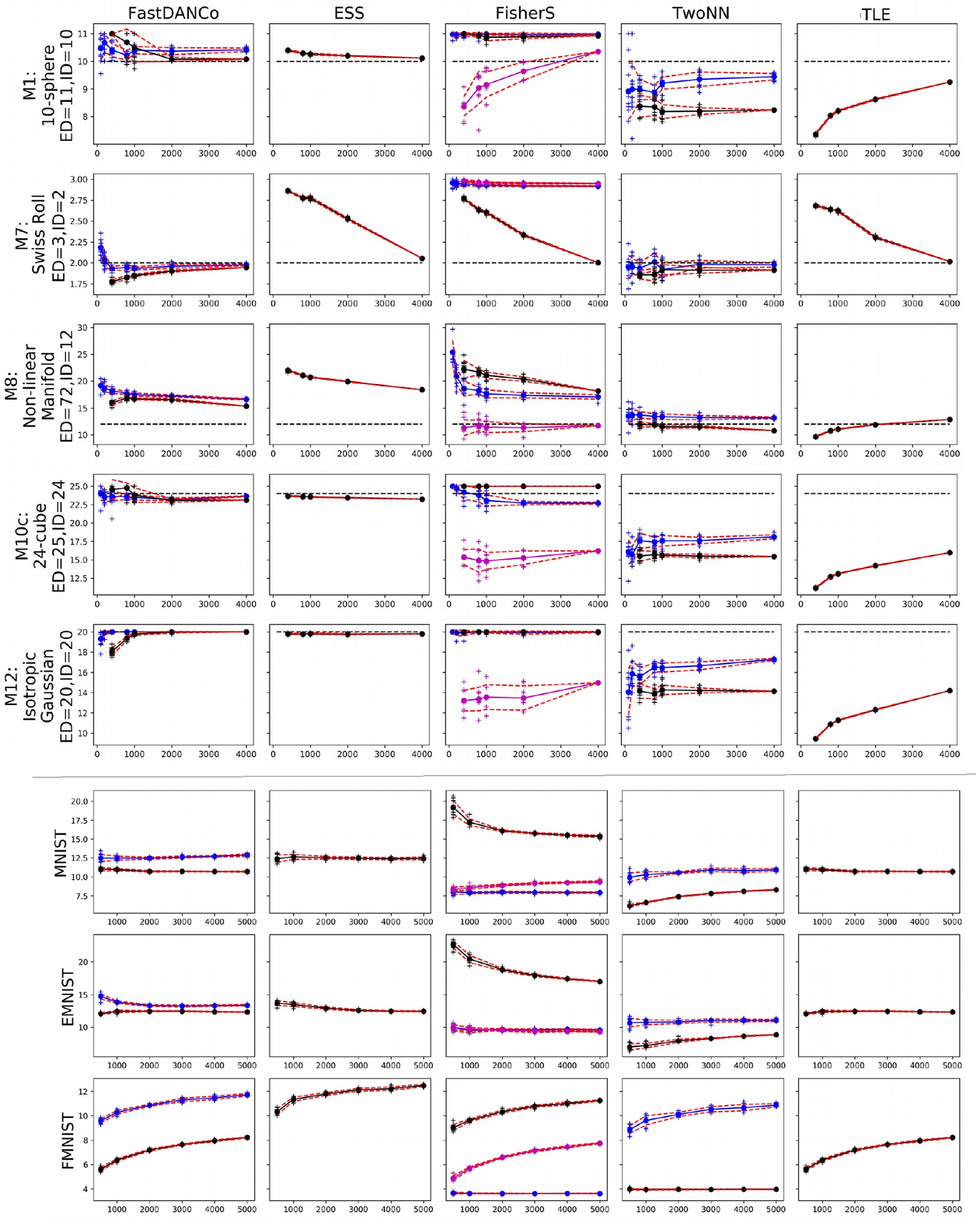}
\caption{Dependence of four ID estimators on subsample size. A selection of synthetic and real-life datasets used in this study is shown. The blue line shows the global ID estimate, while the black one shows the mean local ID. Red dashed lines show confidence intervals. In case of FisherS, the magenta line shows the mean global pointwise ID estimation. For the synthetic datasets, the dashed black line represents true ID. ESS and TLE are local estimators providing no specific procedure to produce global estimates, thus we only show mean local ID.}
\label{fig5}}
\end{figure*}

The results of this analysis are shown in Figure~\ref{fig5}. Notice that for the uniformly sampled $S^{10}$ sphere FisherS determines 11 intrinsic dimensions (as we noticed earlier, FisherS does not distinguish a ball and a sphere). For other uniformly sampled distributions (hypercube, multivariate Gaussian), 3 out of 5 estimators provide consistent results but TwoNN and TLE significantly underestimate the dimension. For the Swiss Roll, global estimators (DANCo and TwoNN) work well while ESS and FisherS show strong dependence of the mean local ID on the size of the subsample. This is expected since forming $k$-nearest neighborhoods with $k=100$ will not respect geodesic distances for lower sample sizes. On the other hand the mean pointwise global estimate is usually biased towards smaller ID. 

For all estimators the dependence of the local ID on the sample size can be quite strong. Overall, it seems that DANCo gives ID estimates closer to TwoNN while ESS matches better FisherS.




\section{Conclusion}

We introduced two new local intrinsic dimensionality (ID) estimators, based on quantifying the separability probability of data points and comparing this statistic to a reference uniform distribution on a unit $n$-dimensional sphere. We performed comparative benchmarking of a number of local ID estimators based on concentration of measure using synthetic and real-life datasets. These benchmarks assessed the ability of the estimators to compute the correct ID from a small sample, their stability to random subsampling and the consistency of their local estimates.

We can conclude that the recently introduced family of local ID estimators exploiting various aspects of concentration of measure, including ours based on linear Fisher separability, performs well on synthetic and real datasets. However, we can distinguish three estimators, FisherS, ESS and DANCo, which seem to be the most adapted to estimate a wide range of ID (from three to several tens) for a neighborhood size of a hundred points or more. ESS showed high agreement with FisherS and can give better results in synthetic cases such as uniform distribution of points in an $n$-dimensional ball with large $n$, while FisherS can give closer to expected estimates for real-life datasets such as ISOMAP Faces. In addition, the global pointwise variant of FisherS estimate combines information about both local and global data point cloud properties, since it quantifies the separability of a data point from all other points, and not only from the closest neighbours. This property might appear useful in some specific applications.

\bibliographystyle{IEEEtran}
\bibliography{IJCNN2020}

\end{document}